\documentclass[10pt,journal,compsoc]{IEEEtran}

\usepackage{multirow}

\ifCLASSINFOpdf
  \usepackage[pdftex]{graphicx}
\else
\fi

\usepackage{amsmath}

\usepackage{algorithmic}

\usepackage{array}

\usepackage{cite}
\usepackage{graphicx}
\usepackage{bm}
\usepackage{caption}
\usepackage{hyperref}
\usepackage{amssymb}
\usepackage{color,xcolor}
\usepackage{subfigure}
\usepackage{algorithm}
\usepackage{algorithm}

\usepackage{ragged2e}

\begin{document}
%
\title{Hybrid Contrastive Learning of Tri-Modal Representation for Multimodal Sentiment Analysis}
%
%
%

\author{Sijie Mai$^1$,
        Ying Zeng$^1$,
        Shuangjia Zheng,
        Haifeng Hu
\IEEEcompsocitemizethanks{\IEEEcompsocthanksitem Haifeng Hu is with School of Electronic and Information Engineering, Sun Yat-sen University,
China.\protect\\
E-mail: huhaif@mail.sysu.edu.cn
\IEEEcompsocthanksitem Sijie Mai, Ying Zeng and Shuangjia Zheng are with School of Electronic and Information Engineering, Sun Yat-sen University,
China.}
\thanks{ }}

\IEEEtitleabstractindextext{%
\begin{abstract}
\justifying
The wide application of smart devices enables the availability of multimodal data, which can be utilized in many tasks. In the field of multimodal sentiment analysis (MSA), most previous works focus on exploring intra- and inter-modal interactions. However, training a network with cross-modal information (language, visual, audio) is still challenging due to the modality gap, and existing methods still cannot ensure to sufficiently learn intra-/inter-modal dynamics. Besides, while learning dynamics within each sample draws great attention, the learning of inter-class relationships is neglected. Moreover, the size of datasets limits the generalization ability of existing methods. To address the afore-mentioned issues, we propose a novel framework HyCon for hybrid contrastive learning of tri-modal representation. Specifically, we simultaneously perform intra-/inter-modal contrastive learning and semi-contrastive learning (that is why we call it hybrid contrastive learning), with which the model can fully explore cross-modal interactions, preserve inter-class relationships and reduce the modality gap. Besides, a refinement term is devised to prevent the model falling into a sub-optimal solution. Moreover, HyCon can naturally generate a large amount of training pairs for better generalization and reduce the negative effect of limited datasets. Extensive experiments on public datasets demonstrate that our proposed method outperforms existing works.
\end{abstract}

}

\maketitle

\let\thefootnote\relax\footnotetext{\textsuperscript{\rm 1}These authors contributed equally to this work}
\IEEEdisplaynontitleabstractindextext

%
\IEEEpeerreviewmaketitle

\section{Introduction}\label{sec:Introduction}

\begin{figure}[h]
  \centering
  \includegraphics[width=0.95\linewidth]{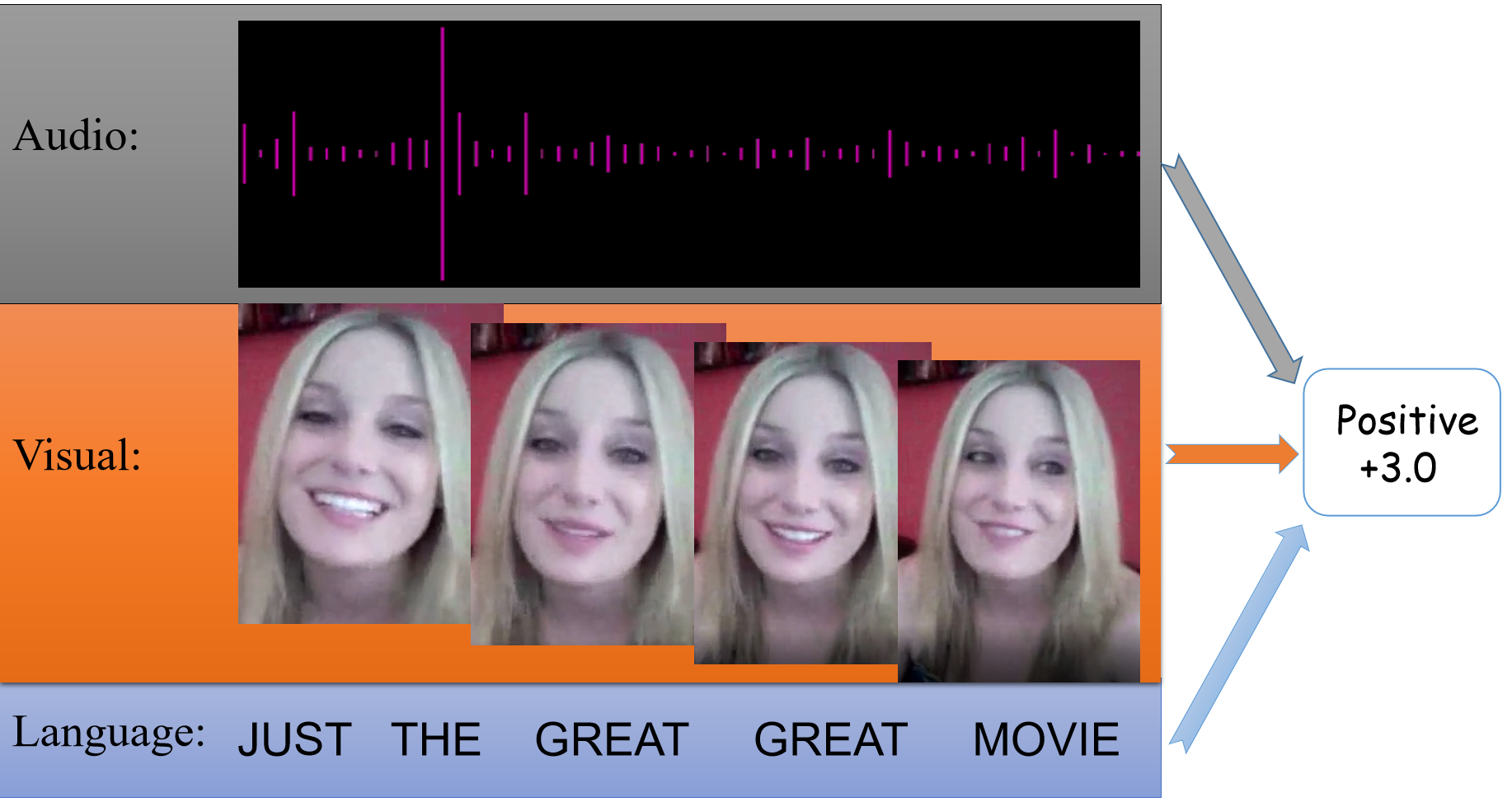}
  \caption{\label{intro}A visualization of multimodal sentiment analysis task. The sample comes from CMU-MOSI dataset \cite{Zadeh2016MOSI}. We aim to predict the sentiment score of a given utterance.}
\vspace{-0.1cm}
\end{figure}

Thanks to the wide applications of smart devices, we can take advantage of abundant multimodal data to perform many downstream tasks \cite{8269806}. As one of the main research directions of multimodal machine learning, multimodal sentiment analysis (MSA) aims to predict the sentiment score from audio, visual, and language features (see Fig.~\ref{intro}). MSA draws increasing attention with the availability of multiple sources of information, and richer information should help to boost performance. However, it has been a challenge to learn meaningful representations for multi-modal data due to the modality gap. Researchers endeavor to devise effective models including RNN variants \cite{Hochreiter1997Long,Cho2014Learning,Zadeh2018Memory,HFFN,Zadeh2018Multi,LMFN,MFRM}, Transformers \cite{transformer,MULT,MRM}, BERT-based models \cite{MAG-BERT,MISA,CM-BERT} to eliminate the gap and learn sufficient interactions between modalities. Those methods mostly focus on learning joint representations in a common manifold, using fusion methods to obtain cross-modal interactions for better performance.

However, most previous works cannot ensure to align modalities well with each other and still suffer modality gap \cite{ARGF}. Those works mostly focus on learning interactions on an intra-class setting, or more specifically, intra-sample setting, which neglect the inter-sample and even inter-class relationships. In other words, prior works seek to learn intra- and inter-modal dynamics within each sample, but not between samples, especially between those samples belonging to different classes. Besides, there exists a tendency of overfitting due to the small size of public datasets and the large amount of parameters introduced by sophisticated designed fusion networks, which degrades the generalization performance of existing MSA models.

To leverage inter-class relationships for better performance, we hypothesize that regularizing distance between different classes and different samples to preserve inter-class relationships help improve the quality of joint cross-modal embedding space, which will be more discriminative for classification/regression and of better generalization ability. A possible solution to implement this hypothesis is to utilize contrastive learning \cite{selfcon1}, which allows the model to sufficiently learn intra-/inter-class cross-modal dynamics. In recent years, a resurgence of work in contrastive learning has led to major advances in representation learning \cite{selfcon1,selfcon2,supcon,COBRA}. Based on it, we propose a novel framework HyCon for hybrid contrastive learning of tri-modal representation. Specifically, HyCon consists of three ways of contrastive learning, i.e., intra-modal contrastive learning (IAMCL), inter-modal contrastive learning (IEMCL) and semi-contrastive learning (SCL), to learn a discriminative cross-modal latent embedding. Unlike prior works that focus on conserving the interactions within each sample, IAMCL and IEMCL allow HyCon to learn intra-/inter-modal dynamics and intra-/inter-class relationships simultaneously. Compared with IAMCL and IEMCL, SCL aims to learn interactions between different modalities within each sample, which can be more focused on aggregating the distribution of different modalities and thus reduce the modality gap.

The technical novelty in this paper is to elaborately devise novel losses based on latest contrastive learning literature \cite{supcon}. For better performance, unlike traditional self-supervised contrastive learning \cite{selfcon1,selfcon2}, our designed IAMCL and IEMCL are performed in a supervised manner. Leveraging label information further ensures that samples from the same class are pulled closer than those from different classes in the feature space. Besides, to reduce modality gap, SCL only considers positive pairs (that is why we call it semi-contrastive learning) to aggregate the modality distribution of the same sample. Moreover, a refinement term and a modality margin are introduced to address the shortcoming of the existing contrastive learning methods \cite{selfcon1,selfcon2,supcon,COBRA} and enable a more suitable representation learning for positive pairs.

As for the problem of limited size of datasets, HyCon takes advantage of contrastive learning to generate a large number of positive and negative pairs for training, with which HyCon can fully learn a more discriminatory boundary between different sentiment classes in the feature space. The inherent advantage of contrastive learning helps to minimize the negative impact of limited datasets and reduce the possibility of overfitting, improving the representation learning ability of the proposed model.

In brief, the main contributions of this paper can be summarized as:
\begin{itemize}
  \item We propose HyCon, a novel framework based on contrastive learning to learn tri-modal representation for MSA which naturally fits the nature of multimodal learning. Considering inter-class relationships that are neglected in existing MSA works, HyCon simultaneously learns cross-modal interactions and explores inter-class relationships to obtain a more sufficient joint embedding.
  \item Three novel ways of contrastive learning are devised to train the model, i.e., IAMCL, IEMCL and SCL. To the best of our knowledge, we are the first to leverage contrastive learning in a hybrid manner to learn cross-modal embeddings. We also introduce a refinement term and a modality margin to enable better representation learning for positive pairs. Learning with our devised losses introduces no additional parameters, so that we can reduce model complexity and the probability of overfitting compared to other methods that introduce high complex module to align different modalities.
\item Our proposed HyCon can leverage a large number of generated sample pairs to acquire better representation learning ability, minimizing the negative impact of limited size of datasets.
 \item Our proposed HyCon is compared with several models on public datasets and achieves state-of-the-art performance even with very simple fusion methods, which demonstrates its effectiveness and superiority. The visualization results suggest that our HyCon can learn a more discriminative embedding space for different classes of sentiments.
\end{itemize}

\section{Related Work}
\subsection{Multimodal Sentiment Analysis}

In recent years, multimodal sentiment analysis (MSA) has attracted significant research interest with the availability of multimodal data \cite{tac_dataset,8269806}. Previous work mostly focus on designing sophisticated fusion strategies to explore inter-modal dynamics\cite{Poria2017A,tac_fusion}.


Two direct ways to perform fusion are \textbf{early fusion}   \cite{Wollmer2013YouTube,Rozgic2012Ensemble,Poria2017Convolutional,Poria2017Context}
and \textbf{late fusion}  \cite{Wu2010Emotion,Nojavanasghari2016Deep,Personality,Zadeh2016MOSI,tac_late_fusion}. Early fusion mainly concatenates the unimodal features at input level to conduct fusion, and late fusion tends to weighted average the unimodal decisions. Those methods outperform unimodal methods, but they can not fully explore intra-/inter-modal dynamics. With the continuous deepening of more advanced fusion methods, \textbf{tensor-based fusion} draws increasing attention for their high expressive power in exploring cross-modal dynamics \cite{Zadeh2017Tensor,HPFN,HFFN}. Specifically, tensor fusion network (TFN) \cite{Zadeh2017Tensor} and Low-rank Modality Fusion (LMF) \cite{Liu2018Efficient} adopt outer product to learn both intra-modality and inter-modality dynamics end-to-end.
Furthermore, some \textbf{modality translation methods} such as Multimodal Transformer \cite{MULT} aims at learning a joint representation by translating source modality into the target one.
Also, \textbf{graph fusion} is considered in Mai et al. \cite{ARGF}, who fuses the modalities using a graph fusion network which regards each interaction as one node. More recently, \textbf{BERT-based methods} such as Multimodal Adaptative Gate BERT (MAG-BERT) \cite{MAG-BERT} and Cross-Modal BERT (CM-BERT) \cite{CM-BERT} achieve significant boost in performance by fine-tuning BERT with cross-modal information.

However, those works are heavily dependent on sophisticated fusion strategies, which introduce more parameters and higher computational costs. Besides, compared to unimodal methods, multimodal methods with complex fusion strategies may easily suffer from overfitting due to the significantly increased number of parameters. The overfitting problem is even more severe when a limited amount of training data is available. More importantly, those methods mostly focus on exploring intra-/inter-modal dynamics to narrow the gap between modalities, but neglect to consider inter-class relationships, which are also informative to help reach more discriminatory boundaries and narrow the modality gap.

Different from them, we investigate the effectiveness of contrastive learning to design a framework HyCon, which can avoid afore-mentioned problems. Our Hycon performs a combination of various ways of contrastive learning with very direct fusion strategy. Despite the simplicity of the fusion method, our proposed model still obtains new state-of-the-art results. To sum up, in addition to considering inter-class relationships for learning more sufficient intra-/inter-modal dynamics, our contrastive learning-based model has the inherent advantage to minimize the impact of limited dataset so as to learn generalized discriminatory features. And it introduces no additional parameters for training, which is time saving and reduce the probability of overfitting.


\begin{figure*}[h]
  \centering
  \includegraphics[width=0.95\linewidth]{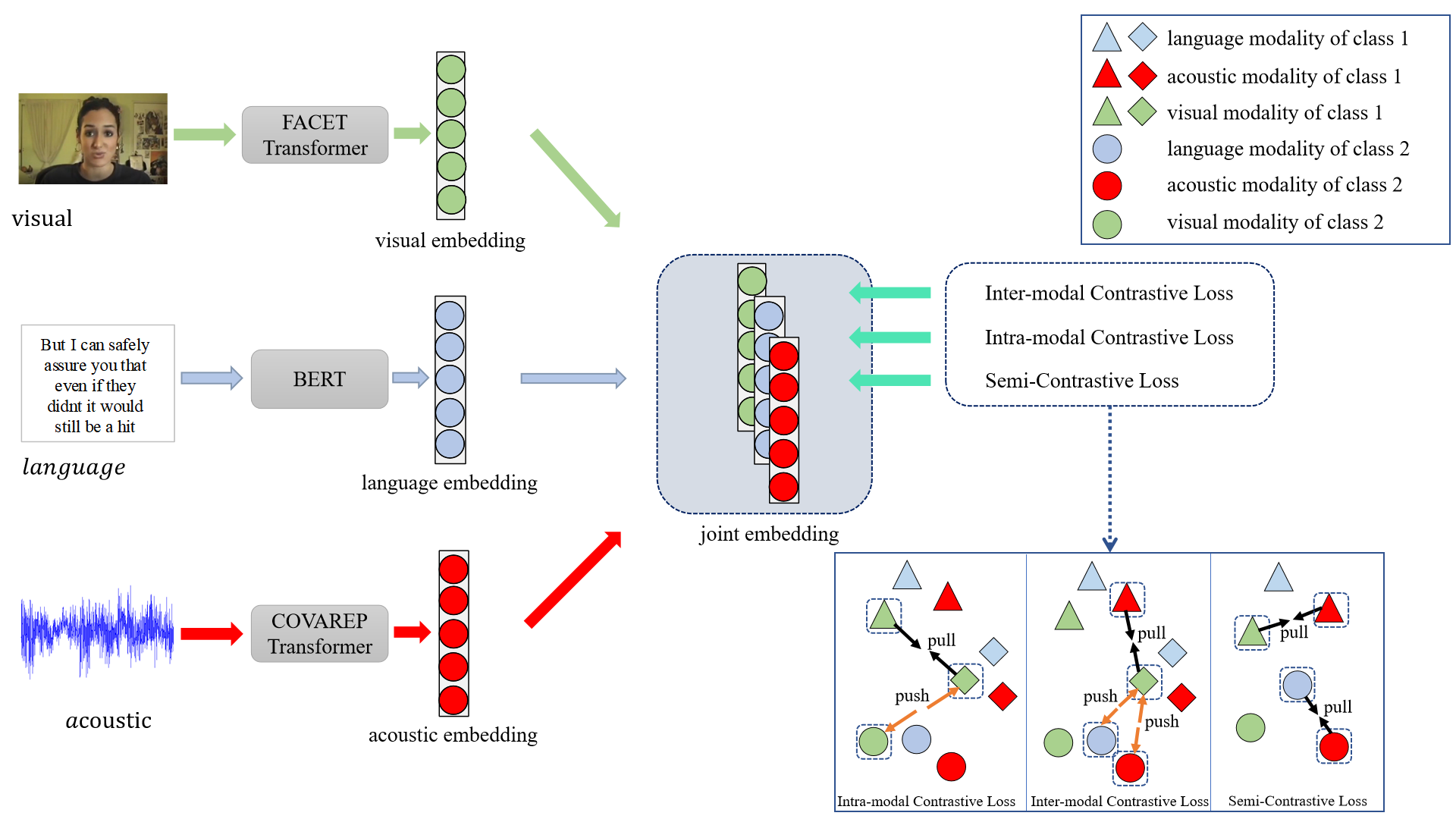}
  \caption{\label{diagram}The diagram of our proposed HyCon and the visualizations of the working of our designed losses.}
\end{figure*}

\subsection{Contrastive Learning}
In recent years, a resurgence of work in contrastive learning has led to major advances in many fields for its effectiveness in learning mutual information of different views of the data. Its principle is quite clear, following the idea that an anchor and a positive sample should be pulled closer in the feature space, while the anchor and negative samples should be pushed apart.

Traditional self-supervised contrastive learning requires one positive sample (e.g., an augmented version of the anchor sample) and many negative samples (e.g., randomly chosen samples from the mini-batch). Self-supervised contrastive learning methods try to learn features from different views, so as to distinguish samples from different classes \cite{selfcon1,selfcon2}. But they only consider one positive sample in each mini-batch and cannot leverage label information even when it is available. Unlike those methods, SupCon \cite{supcon} extends the self-supervised batch contrastive approach to the fully-supervised setting, and designs supervised contrastive loss (SupCon loss). Its positive samples are drawn from samples of the same class as the anchor, rather than being data augmentations of the anchor. SupCon Loss can leverage label information, and the use of many positives and many negatives for each anchor allows us to achieve state-of-the-art performance without the need for hard-negative mining, which can be difficult to tune properly. However, we find that the learning with SupCon Loss may fall into a sub-optimal solution when negative pairs are not sufficient.

Previous works based on contrastive learning mostly only has unimodal data (i.e., images). Contrastive Bi-Modal Representation Algorithm (COBRA) \cite{COBRA} learns joint cross-modal embeddings based on contrastive learning paradigms, which represents the data across different modalities in a common manifold. It achieves this goal by performing supervised contrastive learning, where a positive pair is defined as the representations of data samples belonging to the same modality and class, and a negative pair is defined as the representations of two
samples belonging to same or different modalities of different classes. But COBRA does not utilize many positive samples in each mini-batch, and neglects the inter-modal interactions between different samples from the same class. Thus it may miss out some cross-modal dynamics.

Inspired by the previous works, we propose a novel framework HyCon for multimodal learning. With our hybrid contrastive learning strategy, HyCon can utilize label information for supervised contrastive learning, sufficiently obtaining cross-modal dynamics while at the same time preserve inter-class relationships. We also consider how to avoid the model falling into a sub-optimal solution with a refinement term. Besides, the large amount of sample pairs allows our model to acquire better representation learning ability than existing methods. In a word, our model  can simultaneously preserve intra-/inter-modal interactions and inter-class relationships by constrastive learning in a hybrid manner.

\section{ALGORITHM}

In this section, we describe the proposed method in detail. We first explain the notations and the formulation of the problem statement, after which we introduce the pipeline of the proposed HyCon. The designed loss functions based on contrastive learning are then presented with emphasis. Finally, the optimization and training strategy will be explained.

\subsection{Notations and Problem Formulation}

Our task is to perform multimodal sentiment analysis with multimodal data by scoring the sentiment intensity. The input to the model is an utterance \cite{Olson1977From} (i.e., a segment of a video bounded by pauses and breaths), each of which has three modalities, i.e., audio ($a$), visual ($v$), and language ($l$). The sequences of acoustic, visual, and language modalities are denoted as $\bm{u^a} \in \mathbb{R}^{T_a \times d_a}$, $\bm{u^v} \in \mathbb{R}^{T_v \times d_v}$, and $\bm{u^l} \in \mathbb{R}^{T_l \times d_l}$, where $T_a$, $T_v$ and $T_l$ represent the length of the audio, visual and language modality, respectively, and $d_a$, $d_v$ and $d_l$ denote the dimensionality of the audio, visual and language modality, respectively.

\subsection{Overall Framework}

In this section, we describe the pipeline of our proposed method, with its diagram illustrated in Fig.~\ref{diagram}. Previous works mostly focus on designing fusion strategies to learn cross-modal interactions. However, they only consider intra-class (intra-sample) interactions but neglect the relationships of different samples, especially samples from different classes. Besides, the elaborately designed fusion strategies may introduce high computational costs and may lead to overfitting due to limited size of dataset. To this end, innovated by contrastive learning, we propose a novel framework HyCon to address MSA, which simultaneously considers intra-/inter-modal interactions and intra-/inter-class relationships. HyCon also preserves better representation learning ability with sufficient training pairs.

As shown in Fig.~\ref{diagram}, given an input utterance, we first obtain the unimodal representations via unimodal learning networks. We then generate positive and negative pairs of unimodal representations according to different contrastive losses. After that, in the training stage, the model is trained with our designed contrastive learning to learn cross-modal interactions and inter-class relationships, which will be discarded at inference time. Finally, the learned unimodal embeddings are fused with simple fusion method.

To sum up, HyCon consists of the following main components:
\begin{itemize}
  \item \textbf{Unimodal Learning Network}. To obtain the unimodal latent space of the three modalities, we leverage Transformer \cite{transformer} to extract the features of the audio and visual modalities, and BERT \cite{BERT} to process the language modality of the input utterance, respectively, which can be formulated as:
  \begin{equation}
  \setlength{\abovedisplayskip}{3pt}
\setlength{\belowdisplayskip}{3pt}
\begin{split}
    \bm{X}^{m}&=\bm{F}^m(\bm{u}^m; \theta^m), m \in \{ l, a, v\}\\
  \bm{x}^{m}&= \bm{X}^{m}_{T_m}\in \mathbb{R}^{d}\\
\end{split}
\end{equation}
where $\bm{F}^m$ parametrized by $\theta^m$ refers to each unimodal learning network, $\bm{u}^m\in \mathbb{R}^{T_m \times d_m}$ is the input sequence of modality $m$, and $\bm{x}^{m}$ is the projected unimodal representation.  Note that $\bm{x}^{m}$ is the feature embedding of $\bm{X}^{m}$ in the last time step, and we only use the feature embedding in the last time step to conduct fusion and learning such that our model is suitable for handling unimodal sequences of various length. The feature dimensionality $d$ of unimodal representations is the same across all modalities.

  \item \textbf{Pair Generation}. We construct each mini-batch with $K$ samples (each sample consists of audio, visual, language modalities). During the training stage, the positive pairs and negative pairs are randomly drawn from the mini-batch to train the model according to different losses. Pairs that are several times the number of samples can be generated, so that the model can maximize the use of datasets for better generalization ability.

  \item \textbf{Hybrid Contrastive Learning}. As the core of our proposed method, three versions of contrastive losses operated on the encoded unimodal representations are designed to perform intra-/inter-modal learning in the training stage. With the designed losses, HyCon can sufficiently learn dynamics between and within modalities, preserve inter-class relationships and minimize the modality gap. The designed ways of contrastive learning are as follows:

  1) Semi-Contrastive Learning (SCL): SCL only considers positive pairs to learn interactions between different modalities of the same sample so as to minimize the modality gap.

  2) Intra-modal Contrastive Learning (IAMCL): IAMCL is performed in a supervised manner to learn intra-modal dynamics, considering multiple positive pairs and negative pairs in a mini-batch.

  3) Inter-modal Contrastive Learning (IEMCL): IEMCL is also performed in a supervised manner to learn inter-modal dynamics, and both IAMCL and IEMCL preserve inter-class relationships for more sufficient learning.

  The designed contrastive learning will be explained in detail in Section 3.3.

  \item \textbf{Fusion and Prediction}. With our designed contrastive learning strategy, HyCon can leverage simple fusion method to reach the state-of-the-art performance. In our work, different fusion strategies (including simple and direct ways like concatenation) are adopted to demonstrate the effectiveness and generalization ability of HyCon, which is shown in Section 4.  The defaulted fusion method is `Element-wise Addition'. Generally, the fusion and prediction of our multimodal architecture can be summarized as:
   \begin{equation}
   \label{eq6}
   \setlength{\abovedisplayskip}{3pt}
\setlength{\belowdisplayskip}{3pt}
  y_M = \bm{F}^M(\bm{x}^{l}, \bm{x}^{a}, \bm{x}^{v}; \theta_M), \ell = \left|y-y_M\right|
\end{equation}
where $\bm{F}^M$  parametrized by $\theta^M$ is the multimodal network, $y$ is the ground truth label, $y_M$ is the predicted label, and $\ell$ is mean absolute error (MAE) for prediction.
\end{itemize}

\subsection{HyCon}

The traditional way to perform contrastive learning is self-supervised contrastive learning. A batch of randomly sampled pairs for each anchor consists of a positive pair and $M$ negative pairs. The positive sample is an augmented version of the anchor sample, while each negative sample is an augmented version of the other samples. For each anchor sample, a simple self-supervised contrastive loss can be formulated as:
 \begin{equation}
 \label{con1}
\setlength{\abovedisplayskip}{3pt}
\setlength{\belowdisplayskip}{3pt}
  \bm{L}_{self}=- log \frac {{\bm{a}^T}\bm{p}} {{\bm{a}^T}\bm{p} + \sum_{j=1}^{M}{\bm{a}^T}\bm{n}_{j}}
\end{equation}
where $\bm{a}$, $\bm{p}$ and $\bm{n}$ denote the representation of the anchor, positive and negative sample, respectively.

To leverage label information for better classification performance, SupCon \cite{supcon} extends the traditional self-supervised contrastive learning to the fully-supervised setting. In this setting, each batch is made up of $N$ positive pairs and $M$ negative pairs. A positive sample is any augmented version of samples from the same class, while each negative one comes from different classes. The use of many positives and many negatives for each anchor allows the model to achieve state-of-the-art performance without the need of hard-negative mining. For each anchor sample, supervised contrastive loss can be formulated as:
\begin{equation}
\label{con2}
\setlength{\abovedisplayskip}{3pt}
\setlength{\belowdisplayskip}{3pt}
  \bm{L}_{sup}=- log \frac {\sum_{i=1}^{N}{\bm{a}^T}\bm{p}_{i}} {\sum_{i=1}^{N}({\bm{a}^T}\bm{p}_{i}) + \sum_{j=1}^{M}{\bm{a}^T}\bm{n}_{j}}
\end{equation}

To apply contrastive learning in multimodal data, COBRA \cite{COBRA} adopts similar supervised contrastive loss to learn joint cross-modal embedding. The label information is leveraged in COBRA. For each anchor, the positive sample comes from the same class and the same modality, and the negative sample comes from the same or different modalities of different classes. The equation can be formulated as:
 \begin{equation}
 \label{con3}
\setlength{\abovedisplayskip}{3pt}
\setlength{\belowdisplayskip}{3pt}
  \bm{L}_{COBRA}=- log \frac {{\bm{a}^T}\bm{p}} {{\bm{a}^T}\bm{p} + \sum_{j=1}^{M}{\bm{a}^T}\bm{n}_{j}}
\end{equation}

However, COBRA only considers one positive sample for each anchor as the way in traditional self-supervised contrastive learning. Besides, it only considers positive samples within the same modality but neglect those from different modalities. Consequently, the model may miss out some cross-modal dynamics, which hinders higher performance.

Moreover, observing Eq.~\ref{con1}, ~\ref{con2} and ~\ref{con3}, it can be seen that the learning of existing methods may fall into a sub-optimal solution if $\bm{a}^T\bm{n}_{j}$ is minimized but $\bm{a}^T\bm{p}_{j}$ is not maximized. This may happen when $\sum_{j=1}^{M}{(\bm{a}^m)^T}\bm{n}_{j}$ can be easily learned to be close to 0, especially when sufficient amount of negative pairs is unavailable.

To sufficiently learn cross-modal dynamics and inter-class relationships, while at the same time fully leverages label information, we propose a novel framework HyCon to perform hybrid contrastive learnig in a supervised and unsupervised manner for MSA. In the training stage, hybrid contrastive learning is designed to train the model, which introduces three ways of contrastive learning and their corresponding contrastive losses. And a refinement term is introduced to enable a better representation learning for positive pairs and prevent the model from falling into a sub-optimal solution. We will explain them in detail as follows:

\subsubsection{\textbf{SCL: Semi-Contrastive Learning}}
We first introduce the simplest loss function, i.e., SCL, which is unsupervised.
SCL is devised to learn inter-modal dynamics within the same utterance (sample), so as to minimize modality gap. It only considers positive pairs (that is why we call it semi-contrastive learning), where the positive sample is defined as the representation of data samples from different modalities of the same utterance. For the encoded representation of each modality in each mini-batch $\bm{a}^m$ (i.e., $\bm{x}^m$), it generates two positive samples $S=\left\{\bm{p}_1^{m_1}, \bm{p}_2^{m_2} \right\}$ ($m, m_1, m_2 \in \{l,a,v\}$, $m \neq m_1$, $m_1 \neq m_2$ and $m \neq m_2$).  The scoring function of each pair is based on the common dot product of the representations output by the unimodal learning network.  Firstly, for each modality in each sample, we perform $L2$-normalization on the representation such that the dot product of each pair is between 0 and 1. The semi-contrastive loss can then be formulated as:
\begin{equation}
\setlength{\abovedisplayskip}{3pt}
\setlength{\belowdisplayskip}{3pt}
  \bm{L}_{SCL}= E_s \left[\frac{1}{2}\sum_{i=1}^{2}\left|\left|{(\bm{a}^m)^T}\bm{p}^{m_i}_{i}-\alpha \right|\right|^2\right], \ m \in \{l,a,v\}
\end{equation}
where $E$ is an expectation operator over all the possible sets $S$ in a mini-batch, and $\alpha$ is the modality margin between different modalities which allows for certain modality distributional discrepancies to retain the modality-specific information for fusion.

SCL allows the model to learn the dynamics between modalities of the same sample and draw their embeddings closer in the feature space, in which way the modality gap can be reduced. At the same time, considering that different modalities carry different discriminative modality-specific information that should not be completely eliminated, we allow for certain gap with the set of modality margin $\alpha$. Previous methods tend to explicitly match (translate) the representations of different modalities \cite{ARGF,MULT}, which is rather unrealistic. We argue that different modalities contain discrepancy modality-specific information, it is undesirable and also very difficult to map the representations from different modalities to the same one, which may lead to the loss of unimodal information. Therefore, we let the multimodal fusion stage to learn a richer multimodal representation that fuses information from various modalities rather than directly matching the representation of different modalities.

\subsubsection{\textbf{IAMCL: Intra-modal Contrastive Learning}}

IAMCL aims to learn intra-modal dynamics and inter-class relationships for more discriminative boundaries in the feature space in a supervised manner. In IAMCL, a positive pair is defined as the two unimodal representations from the same modality and the same class of two different data samples, while a negative pair is defined as the unimodal representations from the same modality of two data samples whose classes are different. Considering a mini-batch of size $K$, for each modality $m$ of each anchor in a mini-batch, it generates a set $S=\left\{\bm{p}_1^m, \bm{p}_2^m, ..., \bm{p}_N^m, \bm{n}_1^m, \bm{n}_2^m, ..., \bm{n}_M^m \right\}$ ($m \in \{l,a,v\}$), which consists of $N$ positive samples and $M$ negative samples ($N+M=K-1$). Note that the size of mini-batch is fixed, but the number of $N$ and $M$ is random (i.e., the number of positive and negative pairs in a mini-batch is unfixed).  Similar to SCL, $L2$-normalization on the representation is performed such that the similarity of each pair is between 0 and 1. Then the intra-modal contrastive loss can be formulated as:
 \begin{equation}
\setlength{\abovedisplayskip}{3pt}
\setlength{\belowdisplayskip}{3pt}
\label{eq7}
  \bm{L}_{IAMCL}=-E_s \left[ \frac {\sum_{i=1}^{N}{(\bm{a}^m)^T\bm{p}^m_{i}}} {\sum_{i=1}^{N}{(\bm{a}^m)^T\bm{p}^m_{i}}+\sum_{j=1}^{M}{(\bm{a}^m)^T\bm{n}^m_{j}}}
  \right]
\end{equation}
where $m \in \{l,a,v\}$ denotes the modality $m$, $\bm{a}^m$ denotes the representation of the anchor, $\bm{p}_i^m$ and $\bm{n}_j^m$ represents the representation of positive sample and negative sample, respectively. Observing Eq.~\ref{eq7}, we can notice that the positive and negative samples come from the same modality as the anchor, but their classes are different. Moreover, as demonstrated previously, the learning with traditional supervised loss will be likely to fall into a sub-optimal solution where $\bm{a}^T\bm{n}_{j}$ is minimized but $\bm{a}^T\bm{p}_{j}$ is not maximized. This is because when the similarity of the negative pair approximates zero, the value of the loss is close to 0, regardless of what the similarity of the positive pair is. This phenomena can be more serious when the number of negative pairs is rare, where $\sum_{j=1}^{M}{(\bm{a}^m)^T\bm{n}^m_{j}}$ can be easily learned to be close to 0. In our algorithm, we hope that $(\bm{a}^m)^T\bm{n}^m_{j}$ is minimized and $(\bm{a}^m)^T\bm{p}^m_{j}$ is maximized. Therefore, we introduce a `refinement term' to ensure that the similarity of the positive pairs can be maximized:
 \begin{equation}
\setlength{\abovedisplayskip}{3pt}
\setlength{\belowdisplayskip}{3pt}
\label{eq8}
  \bm{L}^{R}_{IAMCL}=E_s \left[\frac{1}{N}\sum_{i=1}^{N}\left|\left|{(\bm{a}^m)^T}\bm{p}^m_{i}-1\right|\right|^2\right], \ m \in \{l,a,v\}
\end{equation}
 \begin{equation}
\setlength{\abovedisplayskip}{3pt}
\setlength{\belowdisplayskip}{3pt}
\label{eq88}
 \bm{L}_{IAMCL} \gets  \bm{L}_{IAMCL} + \bm{L}^{R}_{IAMCL}
\end{equation}
where $\bm{L}^{R}_{IAMCL}$ is the refinement loss for IAMCL.
IAMCL encourages the the representations of the same modality from different samples but belongs to the same class to have the highest similarity, and forces the representations of the same modality from different classes to have the lowest similarity.  With IAMCL, intra-modal dynamics and inter-class relationships can be learned. With the use of many positive and negative pairs, intra-modal interactions between different samples can be sufficiently preserved without the need of hard-positive and hard-negative learning.

\subsubsection{\textbf{IEMCL: Inter-modal Contrastive Learning}}
IEMCL is similar to IAMCL except that IEMCL aims to learn inter-modal dynamics via contrastive learning, which is neglected in COBRA. Also, IEMCL is different from SCL in that SCL focuses on learning inter-modal interactions within the same sample, but neglects to learn inter-modal interactions between different samples. Specifically, in IEMCL, a positive pair is defined as the two unimodal representations with different modalities from different samples  of the same class, while a negative pair is defined  as the unimodal representations  with different modalities from two samples  whose classes are different. So, for an anchor in a mini-batch of size $K$, compared to IAMCL, it has twice as many negative and positive pairs as the IAMCL. After the $L2$-normalization on all the unimodal representations, the IEMCL loss can be formulated as:
 \begin{equation}
\setlength{\abovedisplayskip}{3pt}
\setlength{\belowdisplayskip}{3pt}
\label{eq9}
      \bm{L}_{IEMCL}=-E_s \left[ \frac {\sum_{i=1}^{2N}{(\bm{a}^m)^T\bm{p}_{i}}} {\sum_{i=1}^{2N}{(\bm{a}^m)^T\bm{p}_{i}}+\sum_{j=1}^{2M}{(\bm{a}^m)^T\bm{n}_{j}}}
  \right]
\end{equation}
where $m \in \{l,a,v\}$ denotes the modality $m$, $\bm{p}_{i}$ and $\bm{n}_{j}$ do not share the same modality as $\bm{a}^m$. Moreover, similar to IAMCL loss, we  introduce a `refinement term' to further stress the learning of the similarity of the positive pairs. And similar to SCL, a modality margin is also introduced to retain modality-specific information for fusion:
 \begin{equation}
\setlength{\abovedisplayskip}{3pt}
\setlength{\belowdisplayskip}{3pt}
\label{eq8}
  \bm{L}^{R}_{IEMCL}=E_s \left[\frac{1}{2N}\sum_{i=1}^{2N}\left|\left|{(\bm{a}^m)^T}\bm{p}_{i}-\alpha \right|\right|^2\right], \ m \in \{l,a,v\}
\end{equation}
 \begin{equation}
\setlength{\abovedisplayskip}{3pt}
\setlength{\belowdisplayskip}{3pt}
\label{eq88}
 \bm{L}_{IEMCL} \gets  \bm{L}_{IEMCL} + \bm{L}^{R}_{IEMCL}
\end{equation}
where $\bm{L}^{R}_{IEMCL}$ is the refinement loss for IEMCL.

\subsection{Training}

The overall contrastive loss function is a weighted sum of IAMCL, IEMCL and SCL, which can be formulated as:

 \begin{equation}
\setlength{\abovedisplayskip}{3pt}
\setlength{\belowdisplayskip}{3pt}
  \bm{L}_{hybrid}= {\lambda_1}{L_{IAMCL}} + {\lambda_2}{L_{IEMCL}} + {\lambda_3}{L_{SCL}}
\end{equation}
where $\lambda_1$, $\lambda_2$ and $\lambda_3$ are hyperparameters to constrain the contributions of the three contrastive losses. ${L}_{hybrid}$ is summed over all data samples of different modalities in a mini-batch. The overall contrastive loss together withe the prediction loss (see Eq.~\ref{eq6}) constitute the loss function for our model:
 \begin{equation}
\setlength{\abovedisplayskip}{3pt}
\setlength{\belowdisplayskip}{3pt}
  \bm{L}_{overall}= \ell + \bm{L}_{hybrid}
\end{equation}
where $\bm{L}_{overall}$ is the overall loss function, and $\ell$ denotes the mean absolute error of the predicted sentiment and the true sentiment.

\section{EXPERIMENT}

\subsection{Datasets}

In this paper, two of the most commonly used public datasets, i.e, CMU-MOSEI \cite{mosei} and CMU-MOSI \cite{Zadeh2016MOSI} are adopted:

\textbf{CMU-MOSI} is a widely-used dataset for multimodal sentiment analysis, which consists of a collection of 2199 opinion video utterances downloaded from websites. Each opinion video is annotated with a sentiment intensity from on -3 to +3. Following previous works \cite{MAG-BERT,MULT}, we utilize 1,284 utterances for training, 229 utterances for validation, and 686 utterances for testing. \textbf{CMU-MOSEI} is a large dataset of multimodal sentiment analysis and emotion recognition. The dataset consists of over 20k video utterances from more than 1,000 YouTube speakers, covering 250 distinct topics. All the sentences utterance are randomly chosen from various topics and monologue videos, and each utterance is annotated on two views: emotion of six different values, and sentiment in the range [-3,3]. In our work, we use the sentiment label to perform MSA. We use 16,265 utterances for training, 1,869 utterances for validation, and 4,643 utterances for testing, which is the same as the previous works  \cite{MAG-BERT,MULT}.

\subsection{Evaluation Metrics}
In our experiments, the evaluation metrics for CMU-MOSEI  and CMU-MOSI datasets are the same. We applie various metrics to evaluate the performance of each model: 1) Acc7: 7-way accuracy, sentiment score classification; 2) Acc2: binary accuracy, positive or negative; 3) F1 score; 4) MAE: mean absolute error (the lower the better); and 5) Corr: the correlation between the model's prediction and that of humans (the higher the better).



\subsection{Baselines}

1) \textbf{Early Fusion LSTM} (\textbf{EF-LSTM}), which concatenates the input features of different modalities at word-level, and then sends the concatenated features to an LSTM layer followed by a classifier to make prediction; 2) \textbf{Late Fusion LSTM} (\textbf{LF-LSTM}) uses an LSTM network for each modality to extract unimodal features and infer decision, and then combine the unimodal decisions by voting mechanism; 3) \textbf{Recurrent Attended Variation Embedding Network} (\textbf{RAVEN}) \cite{RAVEN}, which models interactions by shifting language representations based on the features of the audio and visual modalities; 4) \textbf{Memory Fusion Network} (\textbf{MFN}) \cite{Zadeh2018Memory}, which proposes delta-attention module and multi-view gated memory network to discover inter-modal interactions;  5) \textbf{Multimodal Transformer} (\textbf{MULT}) \cite{MULT}, which learns multimodal representation by translating source modality into target modality using cross-modal Transformer \cite{transformer}; 6) \textbf{Interpretable Modality Fusion} (\textbf{IMR}) \cite{MRM}, which improves the interpretable  ability of MULT by introducing the multimodal routing mechanism;  7) \textbf{Tensor Fusion Network} (\textbf{TFN}) \cite{Zadeh2017Tensor}, which applies outer product from unimodal embeddings to jointly learn unimodal, bimodal and trimodal interactions;  8) \textbf{Low-rank Modality Fusion} (\textbf{LMF}) \cite{Liu2018Efficient}, which leverages low-rank weight tensors to reduce the complexity of tensor fusion; 9) \textbf{Quantum-inspired Multimodal Fusion} (\textbf{QMF}) \cite{Quantum}, which focuses on the interpretable ability of multimodal fusion by taking inspiration from the quantum theory;  10)  \textbf{Interaction Canonical Correlation Network} (\textbf{ICCN}) \cite{ICCN}, which fuses with language embeddings with visual and audio features respectively to get two bimodal representations. Finally, the bimodal representations are fed to a Canonical Correlation Analysis (CCA) network to generate trimodal representation and make prediction;
11) \textbf{Multimodal Adaption Gate BERT} (\textbf{MAG-BERT}) \cite{MAG-BERT}, which proposes an attachment  called Multimodal Adaptation Gate that enables BERT and XLNet to accept multimodal data during fine-tuning. The feature extraction method of MAG-BERT is the same as that of HyCon, which ensures fair comparison. MAG-BERT is the state-of-the-art algorithm on MSA.

\subsection{Experimental Details}

For each algorithm, following Gkoumas et al. \cite{Gkoumas2021WhatMT}, we first perform fifty-times random grid search on the hyper-parameters to fine-tune the model, and save the hyper-parameter setting that reaches the best performance. After that, we train each model with the best hyper-parameters setting for five times, and the final results are the mean results of the five-time running. Note that since the codes of QMF \cite{Quantum} and ICCN \cite{ICCN} are not available, we directly present the results from their papers.

For CMU-MOSEI dataset, the input dimensionality of language, audio, and visual modality is 768, 74, and 35, respectively. While for CMU-MOSI dataset, the input dimensionality of language, audio, and visual modality is 768, 74, and 47, respectively. For feature extraction,  Facet$^1$ is used for the visual modality to extract a set of features that are composed of facial action units, facial landmarks, head pose, etc. These visual features are extracted from the utterance at the frequency of 30Hz to form a sequence of facial gestures over time. COVAREP \cite{Degottex2014COVAREP} is utilized for extracting the representations of audio modality, which include 12 Mel-frequency cepstral coefficients, pitch tracking, speech polarity, glottal closure instants, spectral envelope, etc. These acoustic features are extracted from the full audio clip of each utterance at the sampling rate of 100Hz to form a sequence that represents variations in the tone of voice across the utterance.

\let\thefootnote\relax\footnotetext{\textsuperscript{\rm 1}iMotions 2017. https://imotions.com/}

We develop our model with the Pytorch framework on GTX1080Ti with CUDA 10.1 and torch version of 1.1.0. Our proposed model is trained using Adam \cite{Kingma2014Adam} optimizer whose learning rate is set to 1e-5. The modality margin $\alpha$ is set to 0.8. For convenience, when generating the positive and negative pairs, we only consider the positive and negative classes of sentiment instead of 7-class fine-grained sentiment. $\lambda_1$, $\lambda_2$, and $\lambda_3$ are all set to 1. The feature dimensionality $d$ is set to 50, and the sequence length $T^m$ is 50 for all modalities.

\begin{table}[t]
\centering
 \caption{ \label{t1}\textbf{ The comparison with baselines on CMU-MOSI.} }
\resizebox{.95\columnwidth}{!}{\begin{tabular}{c|c|c|c|c|c}
 \hline
    & Acc7 & Acc2 & F1 & MAE & Corr \\
 \hline
 EF-LSTM  & 31.6 & 75.8 & 75.6 & 1.053 & 0.613  \\
 LF-LSTM  & 31.6 & 76.4 & 75.4 & 1.037 & 0.620  \\
TFN \cite{Zadeh2017Tensor} & 32.2 & 76.4  & 76.3 & 1.017 & 0.604 \\
 LMF \cite{Liu2018Efficient} & 30.6 & 73.8  & 73.7 & 1.026 & 0.602 \\
 MFN \cite{Zadeh2018Memory} & 32.1 & 78.0  & 76.0 & 1.010 &0.635  \\
RAVEN \cite{RAVEN} & 33.8 & 78.8  & 76.9 & 0.968 & 0.667 \\
 MULT \cite{MULT} & 33.6 & 79.3  & 78.3 & 1.009 & 0.667 \\
 QMF \cite{Quantum} & 35.5 & 79.7  & 79.6 & 0.915 & 0.696 \\
 ICCN \cite{ICCN} & 39.0 &  83.0 & 83.0 & 0.860 & 0.710 \\
 MAG-BERT \cite{MAG-BERT} & 42.9 & 83.5 & 83.5 & 0.790 & 0.769 \\
 \hline
HyCon  & \textbf{46.6} & \textbf{85.2}  & \textbf{85.1} & \textbf{0.713} & \textbf{0.790} \\

 \hline
 \end{tabular}}
 \vspace{-0.3cm}
\end{table}%

\begin{table}[t]
\centering
 \caption{ \label{t2}\textbf{ The comparison with baselines on CMU-MOSEI.} Note that IMR cannot perform regression task so that MAE and Corr are unavailable. }
\resizebox{.95\columnwidth}{!}{\begin{tabular}{c|c|c|c|c|c}
 \hline
    & Acc7 & Acc2 & F1 & MAE & Corr \\
 \hline
 EF-LSTM  & 46.7 & 79.1 & 78.8 & 0.665 & 0.621\\
 LF-LSTM  & 49.1 & 79.4 & 80.0 & 0.625 & 0.655 \\
 TFN \cite{Zadeh2017Tensor} & 49.8 & 79.4  & 79.7 & 0.610 & 0.671 \\
 LMF \cite{Liu2018Efficient} & 50.0 & 80.6  & 81.0 & 0.608 & 0.677 \\
 MFN \cite{Zadeh2018Memory} & 49.1 & 79.6  & 80.6 & 0.618 &0.670 \\
 RAVEN \cite{RAVEN} & 50.2 & 79.0  & 79.4 & 0.605 & 0.680 \\
 MULT \cite{MULT} & 48.2 & 80.2  & 80.5 & 0.638 & 0.659 \\
 IMR \cite{MRM} & 48.7 & 80.6  & 81.0 & - & - \\
 QMF \cite{Quantum} & 47.9 & 80.7  & 79.8 & 0.640 & 0.658 \\
  ICCN \cite{ICCN} & 51.6  &  84.2 & 84.2 & \textbf{0.565} & 0.713 \\
 MAG-BERT \cite{MAG-BERT} & 51.9 & 85.0 & 85.0 & 0.602 & \textbf{0.778} \\
 \hline
HyCon & \textbf{52.8} & \textbf{85.4} & \textbf{85.6} & 0.601 & 0.776  \\
 \hline
 \end{tabular}}
\end{table}%

\subsection{Experimental Results}

\subsubsection{\textbf{Comparison with Baselines}}

In this section, we compare our proposed HyCon with other baselines on two datasets CMU-MOSI \cite{Zadeh2016Multimodal} and CMU-MOSEI \cite{mosei}. As shown in Table~\ref{t1} and ~\ref{t2}, although \textbf{MAG-BERT} outperforms other existing methods and sets up a high baseline, it still can be seen that our proposed HyCon obtains the best performance in most cases. Specifically, on CMU-MOSI dataset, our HyCon achieves the best results on all metrics, and it outperforms MAG-BERT by 3.7\% on Acc7, 1.7\% on Acc2 and 1.6\% on F1 score. On CMU-MOSEI dataset, our proposed method yields 0.9\% improvement on Acc7, and 0.4\% on Acc2 and 0.6\% on F1 score. These results demonstrate the effectiveness of our proposed model, indicating the importance of learning intra-/inter-modal dynamics and inter-class relationships in our HyCon.



\subsubsection{\textbf{Ablation Study}}

In this section, we perform ablation studies to verify the effectiveness of the designed contrastive learning method. And we further investigate the contribution of each component by removing it from the model.

Aiming to verify the effectiveness of the designed contrastive losses, we perform ablation experiment where all \textbf{contrastive losses} are removed (see the case of `W/O Contrast Learning' in Table~\ref{t3}). From the experimental result, it can be seen that removing the whole contrastive learning method significantly degrades the performance. Specifically, performance on Acc7, Acc2 and F1 score has seen a great drop. It is obvious that our proposed contrastive learning method is effective and can greatly boost the performance.

Meanwhile, we design two ablation experiments to investigate the contribution of each component in our proposed HyCon. Firstly, we remove the \textbf{refinement loss} from IAMCL and IEMCL (see the case `W/O Refinement Loss') and the performance of the model has a slight drop. The reason may be that without the extra refinement term in IAMCL and IEMCL, the learning has the probability to fall into a sub-optimal solution where the similarity of positive pairs is not optimized. Fortunately, we can generate a certain amount of negative pairs in each mini-batch, so the degradation is not obvious. The refinement term is still significant as it provides a simple and non-parametric way to improve the performance of contrastive learning, especially when the negative pairs are rare. Secondly, we do not consider the \textbf{modality margin} $\alpha$ in SCL and IEMCL (see the case `W/O Modality Margin $\alpha$' where $\alpha$ is set to 1). The results suggest the necessity to consider $\alpha$ in our loss functions, which may be because modality gap is hard to be completely eliminated, and it is reasonable to allow for retaining the modality-specific information for learning a richer multimodal representation.

Finally, we perform ablation study on \textbf{three contrastive losses}. It can be seen that removing any contrastive loss degrades the performance of HyCon, which indicates their effectiveness. Specifically, the performance drops significantly when the IAMCL loss or IEMCL loss is removed. It indicates that \textbf{the learning of inter-class relationships is the fundamental component that leads to the high performance of our model} (as IAMCL and IEMCL aim to learn intra-/inter-modal inter-class relationships), which should be paid more attention in the research of multimodal learning.


\begin{table}[t]
\centering
 \caption{ \label{t3}\textbf{ Ablation studies on the CMU-MOSI dataset.} }
\resizebox{.95\columnwidth}{!}{\begin{tabular}{c|c|c|c|c|c}
 \hline
    & Acc7 & Acc2 & F1 & MAE & Corr \\
 \hline
 HyCon (W/O Contrast Learning) & 43.2 & 82.7 & 82.8 & 0.769 & 0.774  \\
  HyCon (W/O Refinement Loss) & 45.5 & 84.9 &  84.9 & 0.733  &  0.789 \\
 HyCon (W/O Modality Margin $\alpha$) & 44.2 & 84.6 & 84.5  &  0.739 & 0.785  \\
  HyCon (W/O $\bm{L}_{IAMCL}$) & 45.0 & 83.2 & 83.2 & 0.752 &  0.787 \\
HyCon (W/O $\bm{L}_{IEMCL}$) & 42.8 & 83.7 & 83.4 & 0.768 &  0.760\\
 HyCon (W/O $\bm{L}_{SCL}$) & \textbf{47.0} & 84.0 & 84.0 & 0.748 & 0.776   \\
  \hline
HyCon  & 46.6 & \textbf{85.2}  & \textbf{85.1} & \textbf{0.713} & \textbf{0.790} \\
 \hline
 \end{tabular}}
\end{table}%

\begin{table}[t]
\centering
 \caption{ \label{t4}\textbf{ Discussion on the fusion strategies.} Graph fusion \cite{ARGF} regards each unimodal, bimodal, and trimodal interaction as one node, and explicitly models their relationship in a parameter-friendly way. Tensor fusion \cite{Zadeh2017Tensor} applies outer product to explore interactions, which introduces a large amount of parameters and has high space complexity.
 }
\resizebox{.95\columnwidth}{!}{\begin{tabular}{c|c|c|c|c|c}
 \hline
    & Acc7 & Acc2 & F1 & MAE & Corr \\
 \hline
 Concatenation & 44.8 & 84.3 & 84.1 & 0.762 & 0.786   \\
 Addition (defaulted) & \textbf{46.6} & \textbf{85.2}  & 85.1 & \textbf{0.713} & 0.790 \\
 Tensor Fusion \cite{Zadeh2017Tensor} & 43.7 & 83.5 & 83.5 & 0.763 & 0.791 \\
 Graph Fusion \cite{ARGF} & 46.0 & 85.1 & \textbf{85.3} &0.729  &  \textbf{0.801}  \\
 \hline
 \end{tabular}}
\end{table}%

\begin{table}[t]
\centering
 \caption{ \label{t5}\textbf{ Discussion on the selection of modality margin $\alpha$ on CMU-MOSI.} }
\resizebox{.95\columnwidth}{!}{\begin{tabular}{c|c|c|c|c|c}
 \hline
    & Acc7 & Acc2 & F1 & MAE & Corr \\
 \hline

  HyCon ($\alpha=0.5$) & 42.2 & 83.8 & 83.7  &  0.751 & 0.784  \\
   HyCon ($\alpha=0.7$) & 46.3 & 83.8 & 83.8  &  0.737 & \textbf{0.795}  \\
   HyCon ($\alpha=0.8$)  & \textbf{46.6} & \textbf{85.2}  & \textbf{85.1} & \textbf{0.713} & 0.790 \\
HyCon ($\alpha=0.9$) & 46.1 & 84.2 & 84.3  &  0.747 & 0.783  \\
 \hline
 \end{tabular}}
\end{table}%

\subsubsection{\textbf{Analysis of Generalization Ability}}
We also conduct experiments to verify that HyCon is generalized to be applied with different fusion strategies. Previous works mostly rely on sophisticated fusion methods to sufficiently learn cross-modal dynamics to reach satisfactory results. Unlike them, our HyCon can achieve state-of-the-art performance with simple fusion strategies. As shown in Table~\ref{t4}, \textbf{even with simple and direct fusion methods like concatenation and element-wise addition of unimodal representations, HyCon still outperforms all baselines}. Note that \textbf{all the four HyCon variants reach the state-of-the-art performance compared to baselines}. A conclusion can be reached that our designed hybrid contrastive learning is effective and of satisfactory generalization ability.

Moreover, we can notice that the direct addition fusion performs best, while the tensor fusion performs worst. We argue that it is because by using the cross-modal contrastive learning, the modality gap can be significantly reduced, and the feature dimensionality of unimodal and multimodal representations are forced to have approximately the same distribution, such that direct addition is strong enough to explore the complementary information and interactions between modalities. Instead, by applying outer product to explore interactions, the tensor fusion  \cite{Zadeh2017Tensor} introduces a large amount of parameters and has high complexity, which may introduce noise to the feature distribution and degrade the performance of the multimodal system. Moreover, as a parameter-efficient model, graph fusion \cite{ARGF} regards each unimodal, bimodal, and trimodal interaction as one node, and explicitly models their relationship, which also has high performance.

\subsubsection{\textbf{Discussion on the selection of modality margin $\alpha$.}} Having verified the effectiveness of considering the modality margin $\alpha$, we further carry out experiments to investigate the effect of different selections of its value. Our HyCon achieves the best performance when $\alpha$ is set to be $0.8$. From Table~\ref{t5}, we can see that our model suffers degradation when $\alpha$ is set to be either higher or lower than $0.8$. If $\alpha$ is set to be too low (in the case when $\alpha=0.5/0.7$), there exists large modality gap which hinders higher performance. On the contrary, if $\alpha$ is set to be too high (in the case when $\alpha=0.9$), modality-specific information may be lost to align different modalities. It is obvious that the selection of $\alpha$ is of great significance to ensure an optimal solution.

\begin{figure}[htbp]
 \vspace{-0.3cm}
\centering
\subfigure[Learned by HyCon]{
\begin{minipage}[t]{0.45\linewidth}
\centering
\includegraphics[width=1.5in]{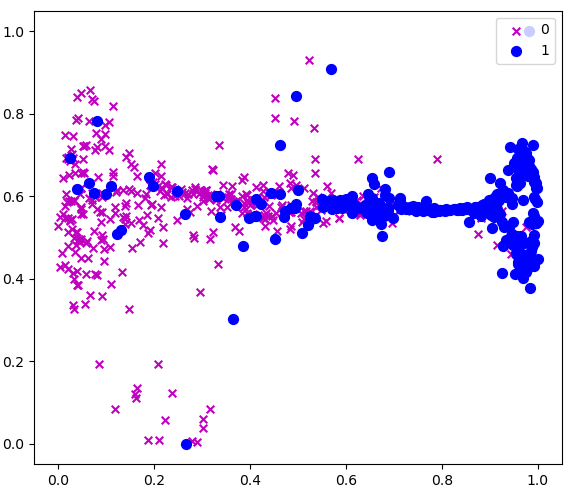}
\vspace{-0.3cm}
\end{minipage}%
 }
 \subfigure[Learned without contrastive learning]{
 \begin{minipage}[t]{0.45\linewidth}
 \centering
 \includegraphics[width=1.5in]{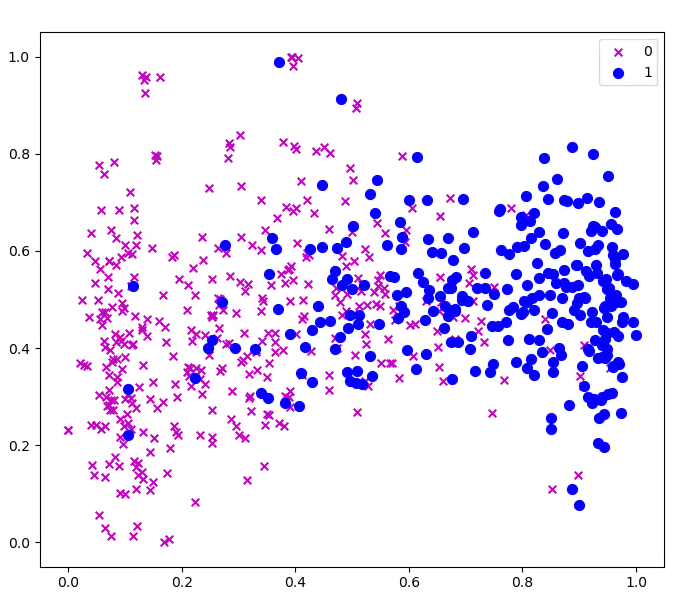}
 \vspace{-0.3cm}
 \end{minipage}
 }%
 \centering
  \vspace{-0.3cm}
 \caption{\label{8}\textbf{T-SNE visualization of the embedding space.} The `red x' and `blue dot' denotes the data point for negative and positive sentiment, respectively.}
 \end{figure}

\subsubsection{\textbf{Visualization for the Embedding Space}}
We provide a visualization for distributions of mutlimodal representation in the embedding space where the left sub-figure on Fig.~\ref{8} illustrates the embedding space learnt by HyCon while the right sub-figure presents the embedding space learnt without contrastive learning. The visualization is obtained by transforming the mutimodal representation into 2-dimension feature point using t-SNE algorithm.  We can infer from Fig.~\ref{8} that when the contrastive losses are removed, the data points in the embedding space tend to be very scattered, and different classes does not form a distinguishing cluster. Instead, \textbf{when the contrastive losses are added to the model, the distance between the data points is significantly narrowed, and each sentiment class form a discriminative cluster.} Moreover, the centers of the two sentiment
clusters are far away, and the data points that are difficult to be clustered are in the middle of the embedding space. This is because we explicitly model the distance (similarity) of different samples, pull the samples of the same class to be closer, and push the samples of different classes to be further apart, which is helpful for the classifier to make prediction.
Nevertheless, there are some dots that are extremely difficult to be correctly classified found in the wrong clusters. This is reasonable because the accuracy of the sentiment classification is only about 85\% even with a well-trained classifier.

\subsubsection{\textbf{Comparison of Loss Functions}}
In this section, we compare the proposed contrastive losses with other loss functions to analyze the effectiveness. The candidate loss functions include the widely-used triplet loss \cite{triplet}, N-pair loss \cite{npair}, and the classical contrastive loss that only considers one positive pair for each anchor. The equations for these losses are presented as follows:
 \begin{equation}
\setlength{\abovedisplayskip}{3pt}
\setlength{\belowdisplayskip}{3pt}
  \bm{L}_{con}=-E_s \left[ \frac {\bm{a}^T\bm{p}} {\bm{a}^T\bm{p}+\sum_{j=1}^{M}{\bm{a}^T\bm{n}_{j}}}
  \right]
\end{equation}
 \begin{equation}
\setlength{\abovedisplayskip}{3pt}
\setlength{\belowdisplayskip}{3pt}
  \bm{L}_{tri}=E_s \left[ ||\bm{a} - \bm{p} ||^2_2 - || \bm{a} - \bm{n} ||^2_2 + 1
  \right]
\end{equation}
 \begin{equation}
\setlength{\abovedisplayskip}{3pt}
\setlength{\belowdisplayskip}{3pt}
  \bm{L}_{n-pair}=E_s \left[ log ( 1+\sum_{j=1}^m exp(\bm{a}^T\bm{n}_{j} - \bm{a}^T\bm{p}) )
  \right]
\end{equation}
where $\bm{a}$ is the anchor, $\bm{n}$ is the negative sample of the anchor, and $\bm{p}$ is the positive sample of the anchor. $\bm{L}_{con}$, $\bm{L}_{tri}$, and $\bm{L}_{n-pair}$ denotes the classical contrastive loss, triplet loss, and N-pair loss, respectively.

We use the above-mentioned losses to replace the proposed $\bm{L}_{IAMCL}$ and $\bm{L}_{IEMCL}$, and the setting of the negative and positive pairs remain the same. However, triplet loss only considers one positive pair and one negative pair where the pairs are randomly sampled from the mini-batch. N-pair loss and the classical contrastive loss consider one positive pair and many negative pairs where the positive pair is randomly sample from the mini-batch and the negative pairs are the same as those of $\bm{L}_{IAMCL}$ and $\bm{L}_{IEMCL}$. One advantage of the supervised contrastive learning is that it can leverage many positive and negative samples for each anchor, which avoids the need of hard-negative mining. We also compare the proposed losses with hard-triplet mining, which has the same equation with the classical triplet loss, but considers hard positive and hard negative pairs. The chosen hard positive sample shares the lowest similarity of anchor among all positive samples in the mini-batch and the hard negative sample shares the highest similarity of anchor.

From Table~\ref{t444}, we can infer that the hard-triplet mining loss performs better than the classical triplet loss significantly on Acc7, Acc2, F1 score and MAE, and reach a similar performance on Corr metric. These results suggest the effectiveness of hard negative/positive mining for triplet loss. However, our HyCon still outperforms hard-triplet mining by a significant margin on all the metrics, mainly for the reason that HyCon considers many positive and many negative pairs for training which is more favorable than one hard negative/positive pair. As for N-pair loss and classical contrastive loss, they both reach satisfactory performance and outperform the classical triplet loss, possibly for the reason that they consider multiple negative pairs. Nevertheless, HyCon performs better than them by a large margin for it generates multiple positive pairs for each anchor. Moreover, HyCon considers modality margin to learn cross-modal mapping, which allows the existence of modality-specific information of each modality for fusion and thereby achieves more favorable performance.

Note that by applying any one of these loss functions, our method can outperform the model without contrastive losses (see Table~\ref{t3} for the results of model without contrastive learning), further demonstrating that the idea of exploring cross-modal and modality-specific dynamics between both intra- and inter-class samples is effective.


\begin{table}[t]
\centering
 \caption{ \label{t444}\textbf{ Comparison of loss functions.} }
\resizebox{.95\columnwidth}{!}{\begin{tabular}{c|c|c|c|c|c}
 \hline
    & Acc7 & Acc2 & F1 & MAE & Corr \\
 \hline
 Triplet Loss  & 43.4 & 83.5  & 83.4 & 0.797 & 0.777 \\
 Hard-triplet Mining  & 45.5 & 84.3  & 84.2 & 0.753 & 0.776 \\
 N-pair Loss  & 45.7 & 83.8  & 83.9 & 0.767 &  0.768 \\
  Classical Contrastive Loss  & 46.3 &  83.7 & 83.6 & 0.740 & \textbf{0.791} \\
  \hline
HyCon  & \textbf{46.6} & \textbf{85.2}  & \textbf{85.1} & \textbf{0.713} & 0.790 \\
 \hline
 \end{tabular}}
\end{table}%

\section{CONCLUSION}
We propose a novel framework HyCon for hybrid contrastive learning of cross-modal representations to perform multimodal sentiment analysis, which is capable of learning intra-/inter-modal dynamics while at the same time exploring inter-class relationships. Specifically, our proposed HyCon consists of three ways of contrastive learning, i.e., intra-modal contrastive learning, inter-modal contrastive learning and semi-contrastive learning. In this way, the designed loss function with a refinement term allows the model to sufficiently learn latent embeddings for sentiment prediction in an optimal way. Moreover, learning with our designed loss function introduce no additional parameters so as to reduce the possibility of overfitting and improve the generalization ability. Experiments demonstrate that our HyCon outperforms state-of-the-art methods.


\ifCLASSOPTIONcaptionsoff
  \newpage
\fi

\bibliographystyle{IEEEtran}


%

%

\bibliography{sample-base,sentiment2}




\end{document}